\documentclass[conference]{IEEEtran}
\IEEEoverridecommandlockouts
\usepackage{cite}
\usepackage{amsmath,amssymb,amsfonts}
\usepackage{algorithmic}
\usepackage{graphicx}
\usepackage{textcomp}
\usepackage{xcolor}
\def\BibTeX{{\rm B\kern-.05em{\sc i\kern-.025em b}\kern-.08em
    T\kern-.1667em\lower.7ex\hbox{E}\kern-.125emX}}

\usepackage{subcaption}
\usepackage{hyperref}
\hypersetup{breaklinks=true,colorlinks=true,linkcolor=blue,citecolor=blue}
\usepackage{algorithm}
\usepackage{algorithmic}
\usepackage{booktabs} 
\usepackage{multirow}
\usepackage{multicol}
\usepackage{float}

\newcommand{\vect}[1]{\mathbf{#1}}

\begin{document}

\title{Multi-time Predictions of Wildfire Grid Map using Remote Sensing Local Data }



\author{\IEEEauthorblockN{Hyung-Jin Yoon}
\IEEEauthorblockA{\textit{Mechanical Engineering Department} \\
\textit{University of Nevada, Reno}\\
hyungjiny@unr.edu}
\and
\IEEEauthorblockN{Petros Voulgaris}
\IEEEauthorblockA{\textit{Mechanical Engineering Department} \\
\textit{University of Nevada, Reno}\\
pvoulgaris@unr.edu}
}

\maketitle

\begin{abstract}
Due to recent climate changes, we have seen more frequent and severe wildfires in the United States. Predicting wildfires is critical for natural disaster prevention and mitigation. Advances in technologies in data processing and communication enabled us to access remote sensing data. With the remote sensing data, valuable spatiotemporal statistical models can be created and used for resource management practices. This paper proposes a distributed learning framework that shares local data collected in ten locations in the western USA throughout the local agents. The local agents aim to predict wildfire grid maps one, two, three, and four weeks in advance while online processing the remote sensing data stream. The proposed model has distinct features that address the characteristic need in prediction evaluations, including dynamic online estimation and time-series modeling. Local fire event triggers are not isolated between locations, and there are confounding factors when local data is analyzed due to incomplete state observations. Compared to existing approaches that do not account for incomplete state observation within wildfire time-series data, on average, we can achieve higher prediction performance.   
\end{abstract}


\section{Introduction}
Wildfires can affect our everyday life in terms of health and property damages~\cite{van2017global, kollanus2017mortality}. Over the last years, the severity of wildfires has increased with longer fire duration and more acres burned~\cite{westerling2006warming}. Wildfire is a vital, dynamic force of nature that shapes the composition and structure of ecosystems. Human interaction with fire has forced it to become both a natural resource management tool, and a hazard that needs to be contained. Anthropogenic-induced climate change, human activity, accumulation of fuels, and past forest management techniques are all contributing factors to the increasing number and severity of forest fires \cite{moritz2014learning}. Despite increased investments in firefighting, the probability and potential losses associated with wildfire risks are increasing and emergency responses are insufficient to protect natural resources and communities \cite{finney2021wildland}. This is reflected on a global scale: North America, South America, Australia, and the Mediterranean Basin have all experienced substantial losses in life and property to wildfires \cite{bowman2011human, chapin2008increasing, holz2017southern, walker2019increasing}. In the western United States alone, the 2020 season saw over 2.5 million hectares burned, the largest wildfire season recorded in modern history \cite{higuera2020record}. Identifying areas with high fire susceptibility is crucial to successfully designing fire management plans and allocating resources \cite{dennison2014large}; consequently, robust approaches to accurately estimate the time, location, and extent of future fires are necessary \cite{sakellariou2019determination, alcasena2019towards}.

\begin{figure}[h]
\vskip -0.2in
\begin{center}
\centerline{\includegraphics[width=0.8\columnwidth]{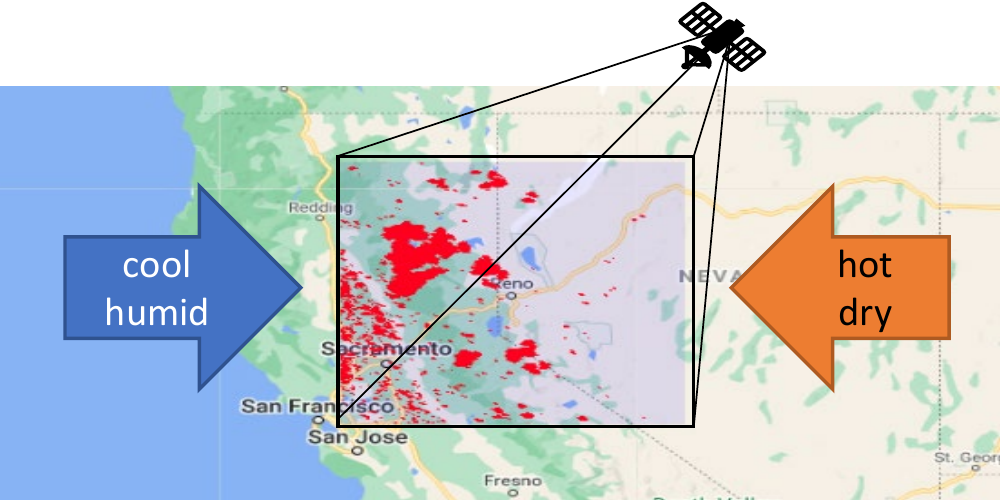}}
\caption{Incomplete state observation of wildfire in local data.}
\label{fig/incomplete_state_observation}
\end{center}
\vskip -0.25in
\end{figure}

The recent advances in data communication technology and computation devices enabled open access to great amount of remote sensing data that can be used for predicting wildfire. For example, \emph{Google Earth Engine} (GEE)~\cite{gorelick2017google} provides time-stamped geographic data in planetary-scale by aggregating the remote sensing data such as \emph{Moderate Resolution Imaging Spectroradiometer} (MODIS)~\cite{giglio2015mod14a1}, \emph{Visible Infrared Imaging Radiometer Suite} (VIIRS)~\cite{vermote2016viirs}, and \emph{Shuttle Radar Topography Mission} (SRTM)~\cite{farr2007shuttle}. With the ease of access to the remote sensing data covering large area utilizing satellite imaginary, data driven approaches have been taken to deal with the challenges in modeling wildfire due to large area that needs to covered with ground sensing data~\cite{jain2020review}. Despite the remarkable success in applications of machine learning (ML), use of ML with the remote sensing data~\cite{gorelick2017google} presents challenges. First, most current ML methods are not able to process planetary-scale satellite images entirely. Second, the majority of ML methods are developed for static data where each samples are collected independently without time-correlation. Lastly, there are confounding factors due to incomplete state observation because the sensor can only measure the output of the planetary system, not necessarily the whole state. There is another source of incomplete state observation when we use local data model due to limited computational resources. Since the neighborhood around the data location influence the local data, the ML prediction  inherently has the issue of incomplete state observation as illustrated in Figure~\ref{fig/incomplete_state_observation}.

\subsection*{Related Works}
There are existing works on wildfire forecasting models. To name an early work, the Canadian Forest Fire Weather Index System uses a mapping designed by domain expert that maps climate features (temperature, humidity, wind, etc) to fire weather index (a fire risk probability)~\cite{van1987development}. Recent literature commonly exhibits spatial fire-susceptibility models using either remotely sensed or agency-reported fire data. An example of this is regional fire-susceptibility mapping in Iran with neuro-fuzzy systems where the predictions of multiple neural networks are aggregated using fuzzy logic~\cite{jaafari2019hybrid}. In~\cite{bashari2016risk}, a Bayesian belief network (hand-crafted by a domain expert) was applied to fire risk prediction. In~\cite{shang2020spatially}, a Random Forest model was used with curated features for fire risk prediction. It is possible to use time-series data with such static model. For example, in~\cite{gholami2021there}, the authors formulated a prediction problem where features includes climate data of the entire twelve month of a year to predict fire risk in the next year. With the curated data, the authors in~\cite{gholami2021there} tested various static models including linear regression, random forest, XGBoost, etc.    

In contrast to estimating a static relation between features to the fire risk~\cite{jaafari2019hybrid, shang2020spatially, gholami2021there, coffield2019machine, mitsopoulos2017data}, dynamic model approaches were also taken. In~\cite{cheng2008integrated}, the authors proposed a spatiotemporal predictive model based on a dynamic recurrent neural network using historical observations in Canada. However, the spatial correlation considered in~\cite{cheng2008integrated} is in the resolution of state, i.e., correlation between states in Canada. A pixel-level spatial-temporal model for fire risk prediction was proposed in~\cite{huot2020deep}. In~\cite{huot2020deep}, the authors aggregated remote sensing data collected using Earth Engine~\cite{gorelick2017google} into a time-series data. The time-series data were fed into an auto-encoder and \emph{Long Short Term Memory} (LSTM) model to estimate fire risk~\cite{huot2020deep}. Furthermore, in~\cite{jin2020ufsp}, \emph{Graph Convolutional Neural Network} (GCNN) was used with the LSTM in order to deal with the high resolution remote sensing data in the fire prediction problem.

Prior literature on wildfire prediction has traditionally and naturally focused on the task, prediction. Majority of the prediction methods used static model~\cite{jaafari2019hybrid, shang2020spatially, gholami2021there, coffield2019machine, mitsopoulos2017data}. The use of time-series models such as reccurent neural networks such as RNN or LSTM has provided more promising results since it can utilize the temporal correlations~\cite{cheng2008integrated, huot2020deep, jin2020ufsp}. However, the aforementioned methods did not consider the incomplete state observation that is common in remote sensing data since the sensor can only indirectly observe the state.In~\cite{yoon2021estimation}, the authors show the effectiveness of a system identification in estimation and decision making when the environment has hidden information. Hence, we proposed to add the system identification task to the prediction task as an auxiliary task.

The auxiliary task is to identify the observation generation model so that the confounding factors due to incomplete state observations can be mitigated. Furthermore, we proposed to share local data between local agents that predict for training of the local agents via another layer of stochastic optimization. We summarize our contributions as follows:
\begin{itemize}
\item We propose to add system identification to predict wildfire in the presence of incomplete state observation.
\item We devise a data sharing scheme between local agents for improved prediction performance while keeping the use of data communication resource under a constraint. 
\end{itemize}

\section{Study Area and Data}
\begin{figure}[t]
\begin{center}
\centerline{\includegraphics[width=0.8\columnwidth]{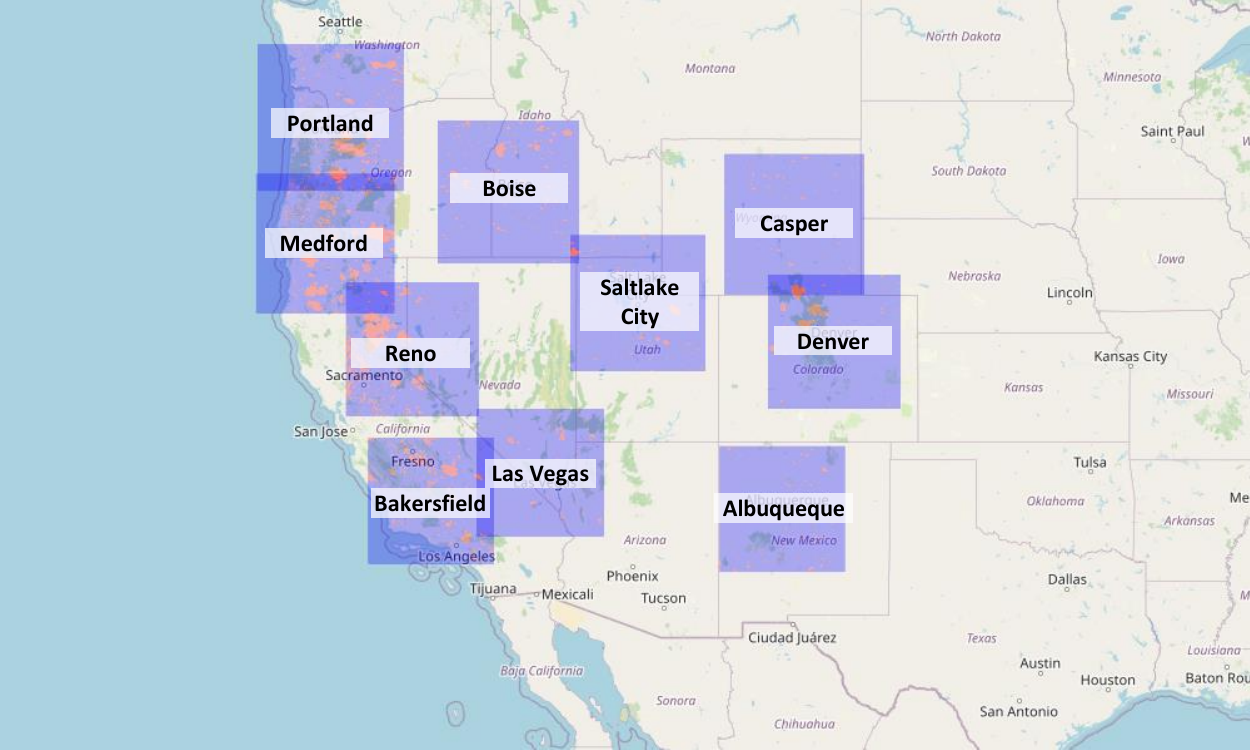}}
\caption{Selected rectangular grids map for data set generation from FIRMS~\cite{davies2019nasa} using \emph{Google Earth Engine}~\cite{gorelick2017google}. }
\label{fig/grid map}
\end{center}
\vskip -0.2in
\end{figure}
In this study, the regions chosen encompass 200 km by 200 km grid squares in the western United States, shown in Figure~\ref{fig/grid map}. The rectangles are centered at cities that have experienced moderate to high wildfire activity in the past decade, including: Portland, Oregon; Medford, Oregon; Reno, Nevada; Denver, Colorado; Bakersfield, California; Boise, Idaho; Salt Lake City, Utah; Casper, Wyoming; Denver, Colorado; and Albuquerque, New Mexico. Within the western United States, the Federal government has designated a higher proportion of land for public access, such as national parks and national forests. Consequently, population growth occurring at wildland-urban interfaces has been introduced to these fire hazards \cite{cannon2009increasing}. These cities, due to their increasing population size, proximity to fire zones, and distribution across the American west, provide a diverse and relevant data set to work with. 

\begin{figure}[h!]
\begin{center}
\centerline{\includegraphics[width=0.5\columnwidth]{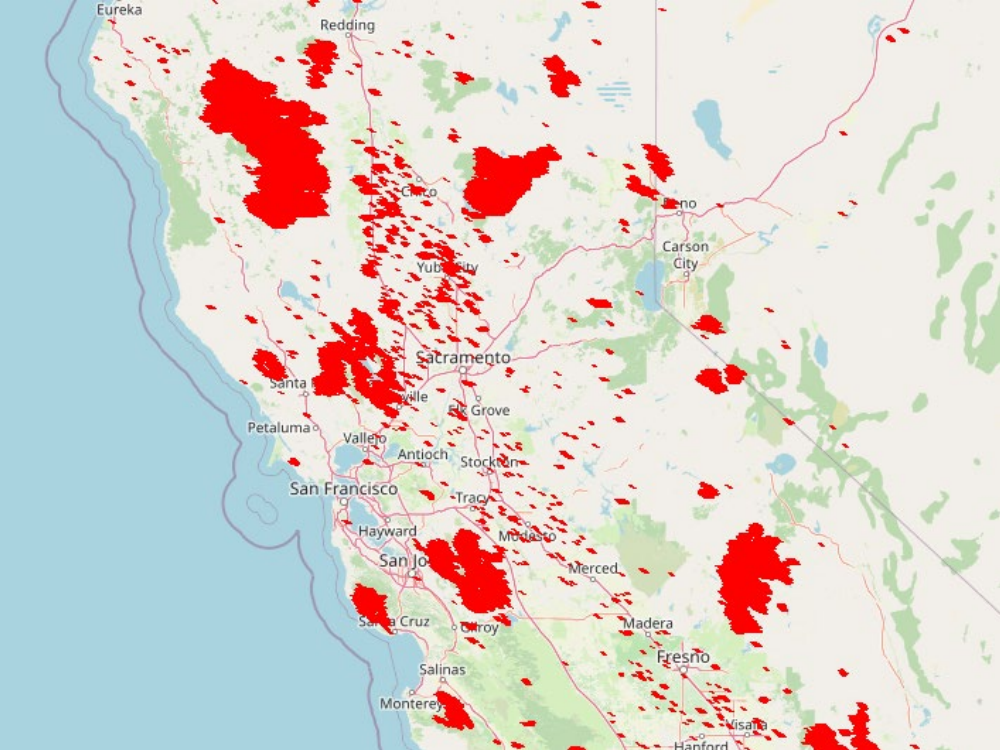}}
\caption{Wildfire in Northern California during Summer 2020.}
\label{fig/firm}
\end{center}
\vskip -0.2in
\end{figure}

\begin{figure}[h!]
\begin{center}
\centerline{\includegraphics[width=\columnwidth]{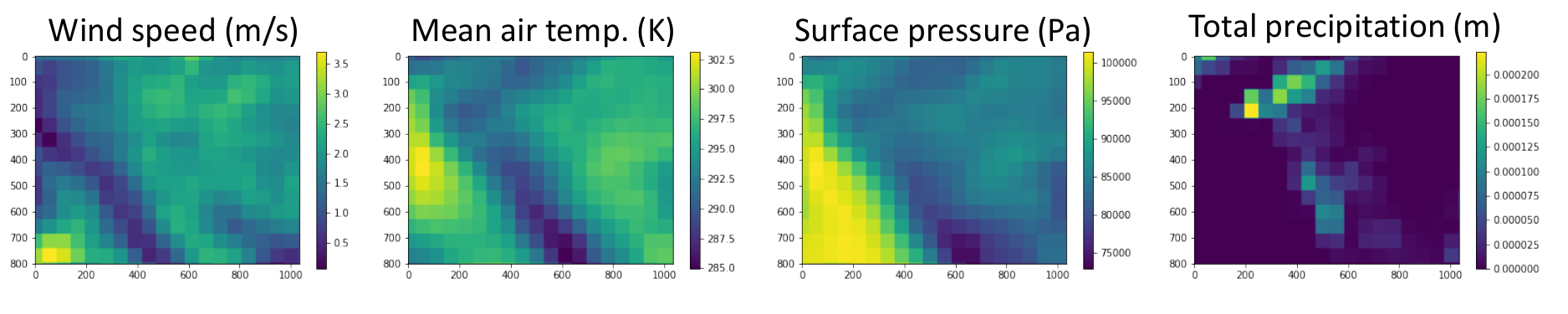}}
\caption{ERA5 Daily Aggregates~\cite{copernicus2017copernicus}}
\label{fig:era5}
\end{center}
\vskip -0.2in
\end{figure}

Data was compiled from multiple remote sensing data sources from Google Earth Engine \cite{gorelick2017google}, a publicly available data repository. We selected sources with high spatial and temporal resolution, extensive geographical and historical coverage, and regular update intervals. Wildfire data is from the FIRMS: Fire Information for Resource Management System data set, containing the LANCE fire detection product in rasterized form that provides near real-time fires detection \cite{davies2019nasa}. Figure~\ref{fig/firm} shows the fire events during Summer 2020 in Western USA, queried from FIRMS.
Climate data is from ERA5 data set produced by Copernicus Climate Change Service, which consists of aggregated values for each day for temperature, dew point, surface pressure, and precipitation, as shown in Figure~\ref{fig:era5} \cite{copernicus2017copernicus}.

\begin{figure}[ht]
\begin{center}
\centerline{\includegraphics[width=\columnwidth]{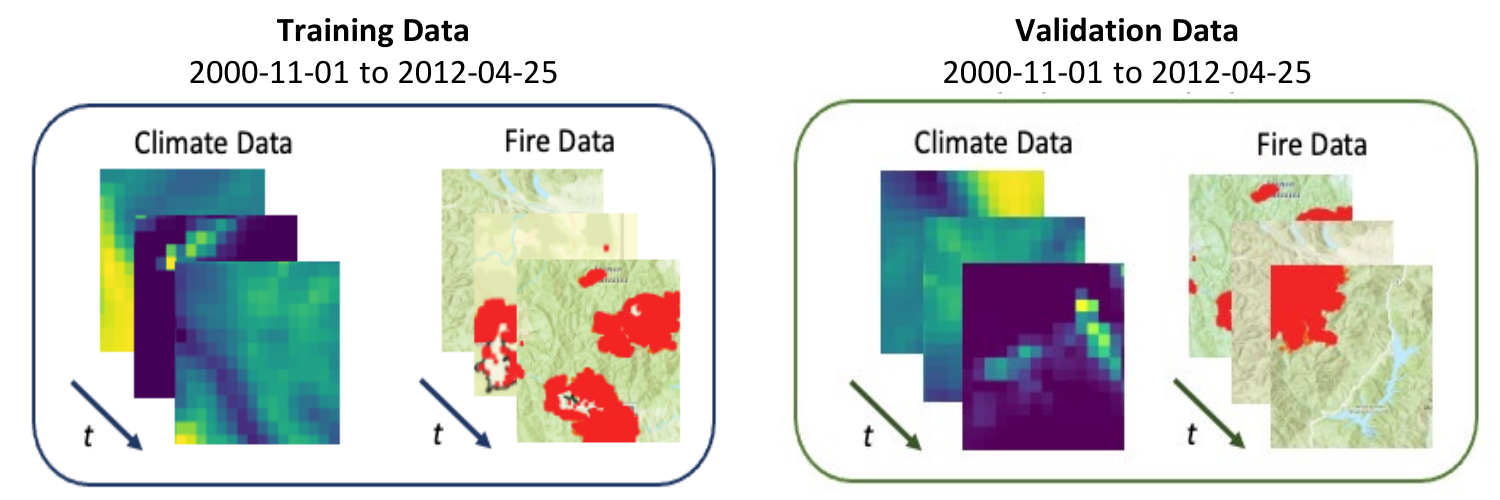}}
\caption{Data partitioning.}
\label{fig/data}
\end{center}
\vskip -0.4in
\end{figure}

The combined data was then partitioned into training and validation data sets, demonstrated in Figure~\ref{fig/data}. The data set is divided temporally in weekly time intervals, \textit{t}, having 800 weeks of training data and 275 weeks of validation data. This data generation scheme holds true for all locations.

\section{Proposed Framework}
\begin{figure*}[h]
\centering
\includegraphics[width=0.95\linewidth]{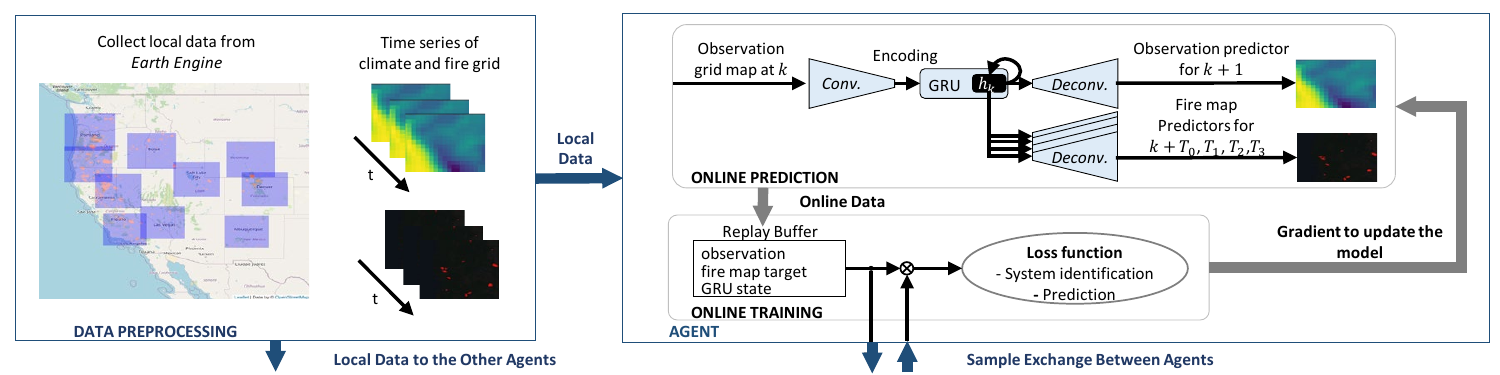}
\caption{Wildfire Prediction Framework with Local Data. }
\label{fig:diagram}
\vskip -0.1in
\end{figure*}
We devise a framework where a local agent process time series of climate and fire grid map online locally while sharing training data between agents for possible performance improvement as illustrated in Figure~\ref{fig:diagram}. The local agent in~\ref{fig:diagram} performs both online prediction and online training. For the online prediction, the agent uses a dynamic computation neural network consists of a set of convolution layers, a gated recurrent unit (GRU)~\cite{chung2014empirical}, and sets of de-convolution layers for one-step prediction of observations and multi-horizon predictions of fire map. The computational network predicts fire map in one, two, three, and four weeks by mapping the state of GRU (detnoted as $h_k$ in Figure~\ref{fig:diagram}) to the grid map through the de-convolution layers.

Training of the computational networks is performed online using data samples from the replay buffer that store local data within the agent including observation, fire map target, and GRU state. In the training, the local agent uses not only the sample from the replay buffer but also samples from the other agents' replay buffers.  There is an high-level decision make that chooses which agent's sample will be suitable to improve performance of the agent. We employed a reinforcement learning for the high-level decision making.

The dynamic computational network performs two types of calculations: (1) the next time step prediction of observation grid map given all previous observations until the current time step while recursively saving the past  observations into $h_k$, state of the GRU; (2) the prediction of fire map in future time (in 1,2,3,and 4 weeks) given $h_k$. The convolutional layers of the computational networks encode the grid map input into low dimensional vectors (encodings) and feeds the encodings to the gated recurrent unit (GRU). Then, the state of the GRU is used by the de-convolution layers to reconstruct the next time step observations and to predict the fire grid maps for the multi-time predictions.

Hence, the loss functions to train the computational network has two terms to minimize. The first term is for the system identification. The reconstruction of the future observation is possible when the computational network can approximate the true dynamics and the observation model of the environment. So, the reconstruction error of the one-time step future prediction of the observation is the first term in the loss function. The second term in the loss function is on its capability in predicting the fire map for multiple future time. Due to the recursive update of the dynamic computational network, we can predict the fire grid map online. As shown in Figure~\ref{fig:diagram}, the dynamic computational network integrates previous observations into the hidden state of GRU. Hence, the model's inference can be implemented in real-time.

The aforementioned loss function terms are for training the local agent using data samples. As shown in Figure~\ref{fig:diagram}, the data samples are not only from the local data but also the local data of the other agents. The exchange of the data sample between the agent is governed by a network policy where we constraint the use of communication resource. We employed another stochastic optimization for improving the network policy. In fact, we implement a reinforcement learning for the stochastic optimization.

To describe the proposed framework in further detail, we will first introduce the overall multi-level optimization consist of the training of the local agents and the reinforcement learning. Then, we will describe the system identification, the prediction, and the reinforcement learning for the network policy in the following sections.

\subsection{Multi-level optimization}\label{sec:multileveloptim}
We use multi-level stochastic optimization that separates the time scales of the updates of the multiple components in Figure~\ref{fig:diagram} and the actor-critic components for reinforcement learning that govern the sample exchanges between agents through a network policy. The learning components are coupled to each other while simultaneously updating the components' parameters. Different choices of the time scales of the update will result in various behaviors in multi-time scale optimization. For example, actor-critic~\cite{konda2000actor} sets a faster update rate for the critic than the actor. And the generative adversarial network in~\cite{heusel2017gans} sets a faster update for the discriminator than the generator. Following the heuristics and theories in~\cite{konda2000actor, heusel2017gans}, we set the slower parameter update rates in lower-level components.

Denote the mapping of the convolution layers in Figure~\ref{fig:diagram} that reduces the dimension of the image into a vector as $\vect{Encoder}(\cdot)$, the gated recurrent unit in Figure~\ref{fig:diagram} as $\vect{GRU}(\cdot)$, and the devolution layers that reconstruct the observation as $\vect{Decoder}_0(\cdot)$. The dynamic computational network for the systems identification is consist of $\vect{Encoder}(\cdot)$, $\vect{GRU}(\cdot)$, and $\vect{Decoder}_0(\cdot)$. We denote the parameter of The dynamic computational network as $\theta_\text{sys}$. We denote the multiple de-convolution layers for fire map prediction in mutiple future times as $\vect{Decoder}_1(\cdot)$ that has the parameter $\theta_\text{pred}$. And the reinforcement learning components for sample exchanges between local agents include the actor denoted as $\mu(\cdot)$ with $\theta_\text{actor}$ and the critic denoted as $q(\cdot \, , \, \cdot)$ that is action-value function for policy evaluation with the parameter $\theta_\text{critic}$.      

The parameters are updated with different step sizes according to the pace of the update rates as follows:
{\small
\begin{equation}\label{eq:multi-time-scale}
  \begin{aligned}
   \begin{matrix}
    \theta^\text{critic}_{n+1}&=&\theta^\text{critic}_{n}&+&\epsilon_n^\text{critic}&
    S_n^\text{critic}(\mathcal{M}_\text{transition})\\
    \theta^\text{actor}_{n+1}&=&\theta^\text{actor}_{n}&+&\epsilon_n^\text{actor}& S_n^\text{actor}(\mathcal{M}_\text{transition})\\
    \theta^\text{sys}_{n+1}&=&\theta^\text{sys}_{n}&+&\epsilon_n^\text{sys}& S_n^\text{sys}(\mathcal{M}_\text{trajectory}, \theta^\text{actor}_{n})\\
    \theta^\text{pred}_{n+1}&=&\theta^\text{pred}_{n}&+&\epsilon_n^\text{pred}& S_n^\text{pred}(\mathcal{M}_\text{trajectory}, \theta^\text{actor}_{n})\\
    \end{matrix}
  \end{aligned}
\end{equation}
}

\noindent where the update functions $S_n^\text{sys}$, $S_n^\text{pred}$, $S_n^\text{actor}$, and $S_n^\text{critic}$ are stochastic gradients with loss functions (to be described in following sections) calculated with data samples from the replay buffers, i.e., $\mathcal{M}_{\text{trajectory}}$ and  $\mathcal{M}_{\text{transition}}$. The replay buffers store finite numbers of recently observed tuples of $(\vect{y}_t, \vect{h}_t)$ and $(\bar{\vect{h}}_{n-1}, \vect{a}_n, r_n, \bar{\vect{h}}_n)$ into $\mathcal{M}_{\text{trajectory}}$ and $\mathcal{M}_{\text{transition}}$ respectively. 

The step-sizes: $\epsilon_n^\text{sys}$, $\epsilon_n^\text{pred}$, $\epsilon_n^\text{actor}$, and $\epsilon_n^\text{critic}$ within the multi-level optimization are set as follows. The predictions of the fire grid map are generated by $\vect{Decoder}_1(\cdot)$ that maps the state of the GRU, $\vect{h}_t$. Hence, we set greater step size for the update of predictor, $\epsilon_n^\text{pred}$, than the step size of dynamic computational network's parameter update, $\epsilon_n^\text{sys}$, that recursively update $\vect{h}_n$. And the update of $\theta^\text{sys}_{n}$ and $\theta^\text{pred}_{n}$ depends of the reinforcement learning components having parameters $ \theta^\text{actor}_{n+1}$ and $ \theta^\text{critic}_{n}$ since the reinforcement learning determines the sample exchange between the agents. Due to the dependence we set slower step size for the reinforcement learning agent. A common practice is that actor-critic~\cite{konda2000actor} sets a greater step size for the critic than the actor because the critic needs to adapt to the change of the actor (policy) for policy evaluation. Hence, the step size follows the diminishing rules as   $n \rightarrow \infty$
{\small
\begin{equation}\label{eq:step-size-rule}
    \frac{\epsilon_n^\text{pred}}{\epsilon_n^\text{sys}} \rightarrow 0 \quad
    \frac{\epsilon_n^\text{sys}}{\epsilon_n^\text{critic}} \rightarrow 0 \quad
    \frac{\epsilon_n^\text{critic}}{\epsilon_n^\text{actor}} \rightarrow 0,
\end{equation}
}

\noindent according to our intention to set slower update rates for lower-level components that generate data for upper-level components.

\subsection{Network Policy Learning via Reinforcement Learning}\label{sec:RL}

\begin{figure}[h]
    \begin{center}
    \centerline{\includegraphics[width=0.5\columnwidth]{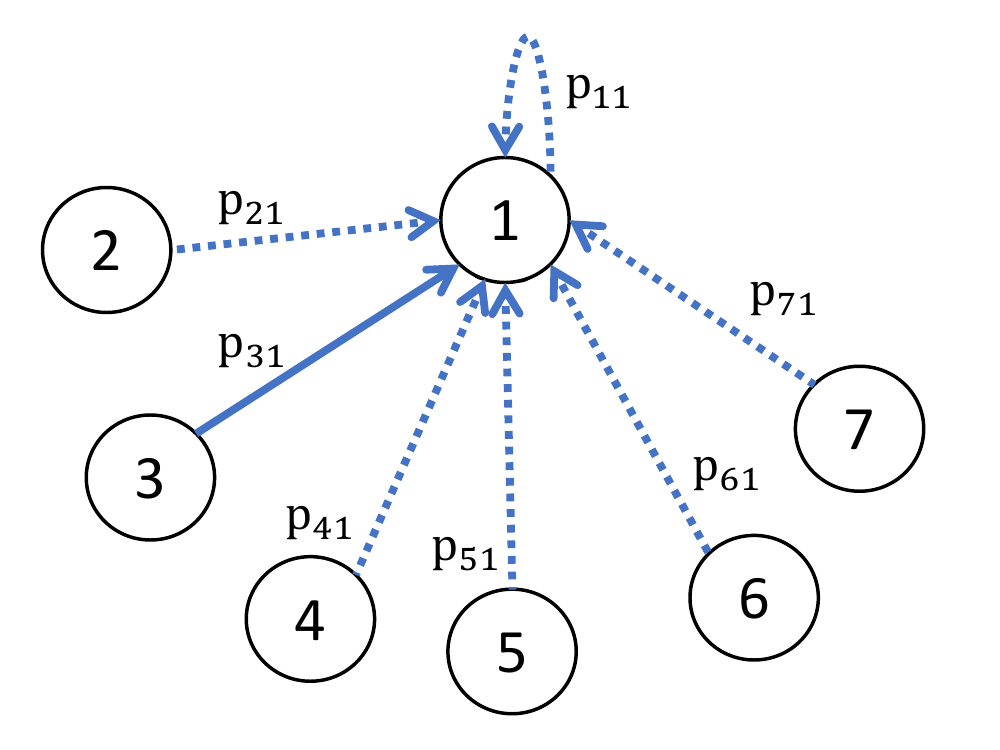}}
    \caption{Network graph describing data sample flow into the $1$ st agent from the other agents. An edge is selected by sampling with the discrete distribution, i.e., $[p_{11}, p_{21}, p_{31}, p_{41}, p_{51}, p_{61}, p_{71}]$.
    }
    \label{fig/network}
    \end{center}
    \vskip -0.3in
    \end{figure}

The sample exchanges between local agents in Figure~\ref{fig:diagram} is selected using the actor of the reinforcement learning. The action by the actor determines the network connection distribution illustrated in Figure~\ref{fig/network}. To regularize the network policy to use the own local data, we set a constraint on the probability to sample from its own data, e.g., $p_{11}=0.8$. The policy determines the distribution as an action, i.e., 
{\small
\begin{equation}\label{eq:policy}
    \vect{a}_n = \mu(\bar{\vect{h}}_n)
\end{equation}
}

\noindent where $\vect{a}_n$ is a matrix having the distribution for the agents as
{\small
\begin{equation}\label{eq:action}
\vect{a}_n := 
\begin{bmatrix}
p_{11}, &p_{21}, &p_{31}, &p_{41}, &p_{51}, &p_{61}, &p_{71}\\
\vdots  &\vdots  &\vdots  &\vdots  &\vdots  &\vdots  &\vdots\\
p_{17}, &p_{27}, &p_{37}, &p_{47}, &p_{57}, &p_{67}, &p_{77}
\end{bmatrix}
\end{equation}
}

\noindent and $\bar{\vect{h}}_n$ is the concatenation of the state of the GRU from the agents as
{\small
\begin{equation}\label{eq:rl_state}
    \bar{\vect{h}}_n:=[\vect{h}_n^{(1)}, \vect{h}_n^{(2)}, \dots, \vect{h}_n^{(7)}]
\end{equation}
}

\noindent where the superscript demotes the index for the local agent. 

The policy by actor, $\mu(\cdot)$, aims to improve the optimization. So, the transition being considered is over the optimization iteration $n$. For the state observation in the optimization process, we chose the state of the GRU sampled from the replay buffer of the $i$ th agent at $n$ as denoted $\vect{h}_n^{(i)}$. To apply a reinforcement learning method, we need to define a reward function. We experimented two reward functions. The first candidate reward function is the value of the intersection over union (IOU) for the online prediction by the local agent. To be specific, the reward value is the average of IOU for the fire grid map by the all local agent. The average is taken over agents and the time horizons considered. So, the objective of the first candidate reward function is to accelerate the training. The second candidate is the maximum value of the loss functions of the agents. Since the RL aims to increase the loss (the adversarial reward) while the optimization algorithm decreases the loss values by selecting more difficult data samples from the other agents. We intended to improve its generalization using the adversarial reward.  

With the action, state observations, and the reward described in the above, we store the state transition (the tuples of previous state estimate, action, reward, and state estimate) in the transition replay memory described as
{\small
\begin{equation}\label{eq:transition_buffer}
\mathcal{M}_{\text{transition}} 
= 
\begin{bmatrix}
(\mathbf{h}_{t-1-L}, &\mathbf{a}_{t-1-L}, &\mathbf{h}_{t-L}, &r_{t-1-L} )\\
 \vdots   &\vdots   &\vdots   &\vdots    \\
(\mathbf{h}_{t-1}, &\mathbf{a}_{t-1}, &\mathbf{h}_{t}, &r_{t-1} )
\end{bmatrix}.
\end{equation}
}

\noindent and optimize the policy using the sample from the replay buffer with the improvement method as follows.

\subsubsection*{Actor-Critic policy improvement}
The critic evaluates the policy relying on the principle of optimality~\cite{bellman1966dynamic}. The use of this principle is helpful to control the variance of the policy gradient by invoking \emph{control variate} method~\cite{law2000simulation}. We employed an actor-critic method~\cite{lillicrap2015continuous} for the reinforcement learning component in the proposed framework. The critic network calculates the future expected rewards given the current state and the action, i.e., $\mathbb{E}\left[\sum \gamma^n r_n| \bar{\mathbf{h}}_0 = \bar{\mathbf{h}}, \vect{a}_0 = \vect{a}\right] \approx q(\bar{\mathbf{h}}, \vect{a};\theta_\text{critic})$ where $\bar{\vect{h}}$ was described in~\eqref{eq:rl_state}.

The critic is updated with the stochastic gradient as
{\small
\begin{equation}\label{eq:critic}
    S^\text{critic}_n = -\nabla_{\theta_\text{critic}} l^\text{critic}(\mathcal{M}_\text{transition};\theta_\text{critic})
\end{equation}
}

\noindent with the following loss function\footnote{We follows the approach of using target networks in~\cite{lillicrap2015continuous} to reduce variance during stochastic updates.}
{\small
\begin{equation}\label{eq:loss_critic}
    l^\text{critic}(\mathcal{M}_{\text{transition}};\theta_\text{critic})
    =\frac{1}{M}\sum_{m=1}^M(q(\bar{\mathbf{h}}_m, \mathbf{a}_m; \theta_\text{critic})- q_m^{\text{target}})^2,
\end{equation}
}

\noindent where $q_m^{\text{target}} = r_m + \gamma q(\bar{\mathbf{h}}'_{m}, \mu_\theta(\bar{\mathbf{h}}'_m);\theta_\text{critic})$ and the state transition samples $\left((\bar{\mathbf{h}}, \mathbf{a}, \bar{\mathbf{h}}', r)_0, \dots, (\mathbf{h}, \mathbf{a}, \mathbf{h}', r)_M\right)$ are sampled from the replay buffer $\mathcal{M}_{\text{transition}}$ in~\eqref{eq:transition_buffer}.
With the same state transition data samples, we calculate the stochastic gradient for the policy using the estimated policy as
{\small
\begin{equation}\label{eq:grad_actor}
\begin{aligned}
    S^\text{actor}_n &= \nabla_{\theta_\text{actor}} J(\mathcal{M}_\text{transition};\theta_\text{actor}), \\
    J &= \frac{1}{M}\sum_{m=1}^M q(\bar{\mathbf{h}}_m, \mu(\bar{\mathbf{h}}_m; \theta_\text{actor});\theta_\text{critic}).
\end{aligned}
\end{equation}
}
\subsection{Fire Map Prediction using Deconvolution Layers}

The deconvolution layers in Figure~\ref{fig:diagram} generate the predicted fire grid map denoted as $\hat{\vect{f}}_t$ given the state of the GRU denoted as $\vect{h}_t$, i.e., $\hat{\vect{f}}_t = \vect{Decoder}_1(\vect{h}_t;\theta_\text{pred})$. $\mathbf{h}_k$ is mapped from all previous observations up to the current time through the recursive update in~\eqref{eq:dynautoenc}. The parameter iterate of the deconvolution layer $\theta^\text{pred}_n$ is updated with the stochastic gradient as
{\small
\begin{equation}\label{eq:pred}
    S^\text{pred}_n =  - \nabla_{\theta_\text{pred}} l^\text{pred}(\mathcal{M}_n^\text{trajectory};\theta_\text{pred}),
\end{equation}
}

\noindent where the loss function $l^\text{pred}(\cdot)$ is calculated as follows:
{\small
\begin{equation*}
    l^\text{pred}(\mathcal{M}_n^\text{trajectory};\theta_\text{pred}) = \frac{1}{LK}\sum_{l=1}^L\sum_{k=1}^K H(\mathbf{f}_{t_k+T_l}, [\hat{\mathbf{f}}_{t_k}]_l)
\end{equation*}
}

\noindent where $\mathbf{f}_{t_k+T_l}$ denotes the ground truth fire grid map at time $t_k+T_l$ and $\hat{\mathbf{f}}_{t_k}$ denotes the mapping of $\vect{h}_{t_k}$ using the deconvolution layer, i.e.,
{\small
\begin{equation}\label{eq:predictor}
    \hat{\mathbf{f}}_{t_k}:=[\hat{\mathbf{f}}_{t_k,1}, \dots \hat{\mathbf{f}}_{t_k,L}] = \text{Decoder}_1(\vect{h}_{t_k};\theta_\text{pred})
\end{equation}
}

\noindent and $\vect{h}_{t_k}$ are sampled from $\mathcal{D}_n^\text{trajectory}$ as a minibatch that is $(\vect{h}_{t_1}, \vect{h}_{t_2}, \dots, \vect{h}_{t_K})$ and corresponding ground truth fire grid maps are queried as $\mathbf{f}_{t_k+T_l}$ for all $t_k$ and $T_l$, and $H(\cdot, \cdot)$ denote an error function to minimize for the prediction. We use the crossentropy function for the error function since the crossentropy is widely used for image segmentation task~\cite{howard2019searching, voigtlaender2019feelvos}. The crossentropy error function is calcualted as
{\small
\begin{equation*}
    H(\vect{f}, \hat{\vect{f}}) = \frac{1}{WH}\sum_{i=1}^{W}\sum_{j=1}^{H} h([\vect{f}]_{i,j}, [\hat{\vect{f}}]_{i,j})
\end{equation*}
}

\noindent where the binary cross entropy is calculated as $h(x, \hat{x})=x\log\hat{x} + (1-x)\log(1-\hat{x})$ for $x\in \{0,1\}$ and $\hat{x}\in (0,1)$, and we follow the convention $0 = 0 \log 0$. The binary label $x\in \{0,1\}$ indicates whether the location of the pixel in the grid map has an active fire, i.e., $x=1$ or not i.e., $x=0$.

\subsection{System identification for state estimation}\label{sec:state_estimation_learning}
The system identification aims to determine the parameter that maximizes the state estimate's likelihood. We maximize the likelihood of state predictor by minimizing the cross-entropy error between true image streams and the predicted image streams by a stochastic optimization which samples trajectories saved in the memory buffer denoted by $\mathcal{M}_\text{trajectory}$ with a loss function to minimize. We use replay buffer~\cite{zhang2017deeper} to save recent trajectories to sample minibatch samples for training as follows:
{\small
\begin{equation*}
\mathcal{M}_{\text{trajectory}} =
\begin{bmatrix}
(\mathbf{x}_{0}, \mathbf{h}_{0})_1, &\dots, &(\mathbf{x}_\text{term}, \mathbf{h}_\text{term})_1\\
 \vdots & \vdots & \vdots    \\
(\mathbf{x}_{0}, \mathbf{h}_{0})_{N}, &\dots, &(\mathbf{x}_\text{term}, \mathbf{h}_\text{term})_{N}
\end{bmatrix}
\end{equation*}
}

\noindent where $\mathbf{y}_k$ denotes the observations of the map at time $k$ that are extracted from the climate~\cite{copernicus2017copernicus} and fire data set\cite{davies2019nasa}.   

The loss function $l^\text{sys}(\cdot)$ is calculated using the sampled trajectories from $\mathcal{M}_{\text{trajectory}}$ 
We calculate the loss function as
{\small
\begin{equation}\label{eq:loss_sys_id}
    l^\text{sys}(\mathcal{M}_{\text{trajectory}};\theta_\text{sys}) = \frac{1}{M}\sum_{m=1}^M d(\mathbf{X}_m, \hat{\mathbf{X}}_m)
\end{equation}
}

\noindent where $\mathbf{X}_m = (\mathbf{x}_0, \dots, \mathbf{x}_T)_m$ is the $m$\textsuperscript{th} sample image stream with time length $T$. Here, $H(\cdot, \cdot)$ is average of the binary cross-entropy $h(\cdot, \cdot)$ between the original image stream $\mathbf{X}_m$ and the predicted image stream $\hat{\mathbf{X}}_m$ as
{\small
\begin{equation*}
    d(\mathbf{X}_m, \hat{\mathbf{X}}_m)
    = \frac{1}{TWHC}\sum_{t=1}^{T}\sum_{i=1}^{W}\sum_{j=1}^{H}\sum_{k=1}^{C} ([\mathbf{X}_m]_{t,i,j,k} - [\hat{\mathbf{X}}_m]_{t,i,j,k})^2
\end{equation*}
}

\noindent where $i,j,k$ denotes width, height, color index for the image with width $W$, height $H$, number of channel $C$ and we use the mean squared error to calculate the difference between the pixel intensities of $[\mathbf{X}_m]_{t,i,j,k} \in [0,1]$ and   $[\hat{\mathbf{X}}_m]_{t,i,j,k} \in [0,1]$.

We generate the predicted trajectory given the original trajectory with image stream $\mathbf{X}_m = (\mathbf{x}_0, \dots, \mathbf{x}_T)_m$ by processing them through the encoder, GRU, and the decoder as
{\small
\begin{equation}\label{eq:dynautoenc}
    \begin{aligned}
    \mathbf{h}_{t+1} &= \text{GRU}(\mathbf{h}_t, \text{Enc}(\mathbf{x}_t)) \\
    \hat{\mathbf{x}}_{t+1} & = \text{Decoder}_0(\mathbf{h}_{t+1})
    \end{aligned}
\end{equation}
}

\noindent and collect them into $\hat{\mathbf{X}}_m =\{\hat{\mathbf{x}}_1, \dots, \hat{\mathbf{x}}_T\}$. With the loss function in~\eqref{eq:loss_sys_id}, the stochastic gradient for the optimization is defined as
{\small
\begin{equation}\label{eq:grad_sys_id}
    S^\text{sys}_n =  -\nabla_{\theta_\text{sys}} l^\text{sys}(\mathcal{M}_\text{trajectory};\theta_\text{sys}).
\end{equation}
}
We summarise the entire procedure as the following multi-level stochastic optimization in Algorithm~\ref{alg:multi-level-optimization} at Appendix.

\section{Evaluation and Discussion }
\begin{figure*}[ht]
\centering
\includegraphics[width=\linewidth]{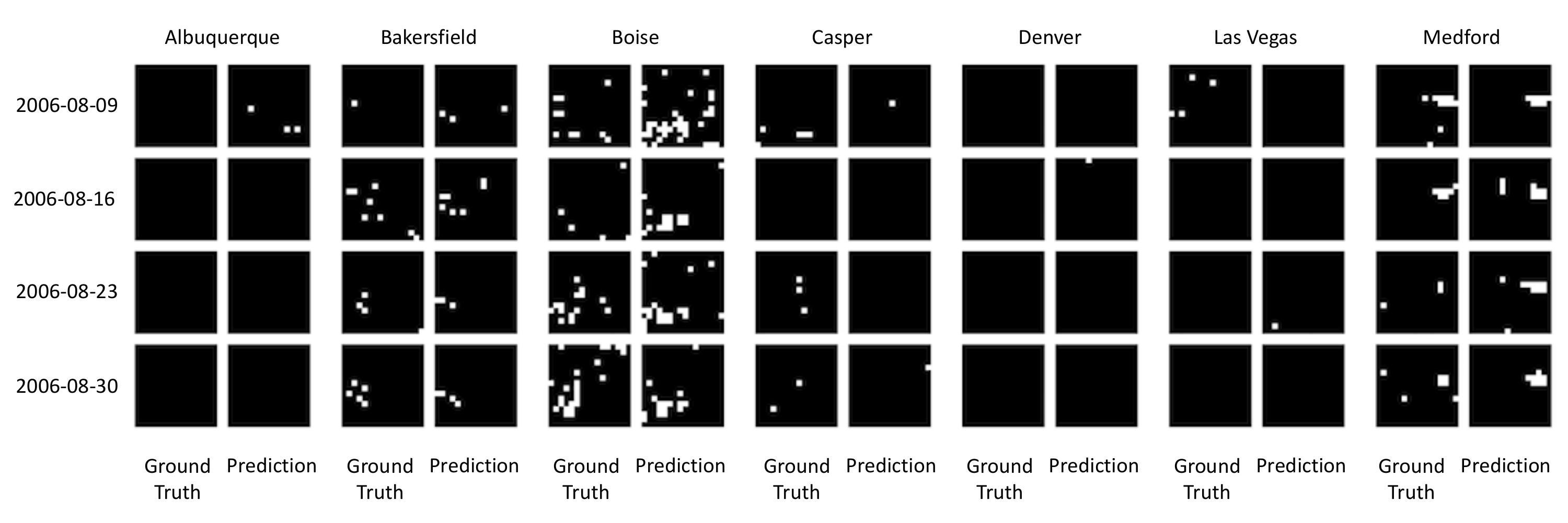}
\caption{Multiple future predictions made on 2006-08-02. See the illustrative video linked at \url{https://youtu.be/Z_Jkr8yzVFA}.}
\label{fig:future-predictions}
\end{figure*}
\begin{figure*}[t]
\centering
\includegraphics[width=0.95\linewidth]{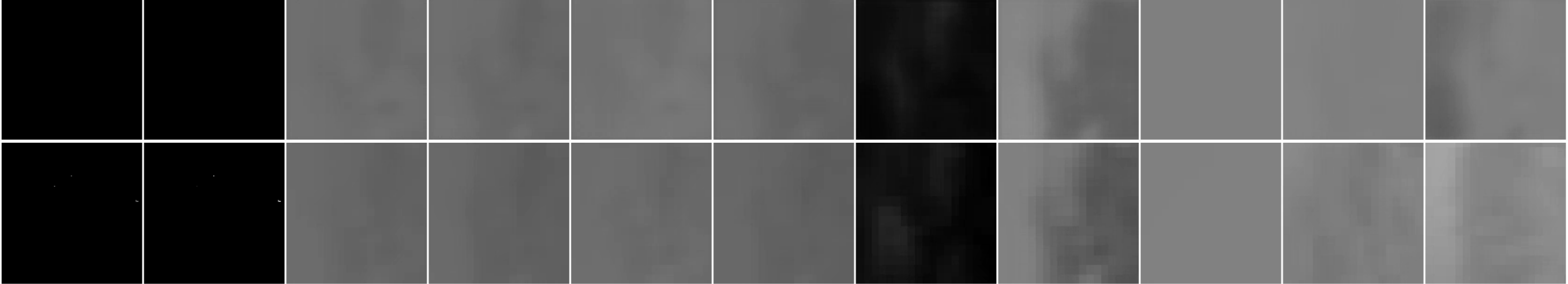}
\caption{One-step future observation prediction for state estimation. The top row shows the prediction and the bottom row shows ground truth. The first two columns are from FIRMS~\cite{davies2019nasa} and the other columns are from ERA5 Daily Aggregates~\cite{copernicus2017copernicus}. 
}
\label{fig:one-step-prediction}
\end{figure*}

We trained the proposed model in Figure~\ref{fig:diagram} using time series data from 2000-11-01 to 2012-04-25 and applied the trained model to validation data from 2012-05-02 to 2016-02-24. As illustrated in Figure~\ref{fig:diagram}, the local agent exchanges sampled data with each other using the policy of a reinforcement learning agent. The training of the local agents and the reinforcement learning agents are simultaneously run as described in Algorithm~\ref{alg:multi-level-optimization}. Since the algorithm process the time-series data online, we train the model using the time-series up to 2012-04-25. And a trained model processes the time-series data up to 2016-02-24 and uses the performance for the duration from 2012-05-02 to 2016-02-24 in order to evaluate the model's performance. The online predictions of the future fire grid map are shown in Figure~\ref{fig:future-predictions}. In Figure~\ref{fig:future-predictions}, The trained model of local agents recursively process the time series up to 2006-08-02 and predict the fire grid map for the future dates in 1, 2, 3, and 4 weeks. We use cross-entropy and Area Under ROC Curve (AUROC) to evaluate the prediction performance. We also report the intersection over union (IOU) as additional information. However, the pixel positions with the same label in the ground truth fire grid map are not clustered as typical image segmentation tasks. So, the value of IOU in this paper is much less than the result in other segmentation task research papers. For example, the average IOU between the ground truth and the prediction in Figure~\ref{fig:future-predictions} is $11\%$, although the average AUROC for the predictions is $98\%$.          

We compared the proposed model with alternatives. The first alternative is a Logistic regression, where a linear model is trained to map the feature values of a single-pixel point into the risk of fire. The second alternative is the generative network (denoted as Generative Network in Table~\ref{tb:comparative_analysis}) is comprised of an $\text{Enc}(\cdot)$ and $\text{Decoder}_1(\cdot)$ from the computational networks in Figure~\ref{fig:diagram}. In the generative network, the dynamic component $\text{GRU}(\cdot, \cdot)$ is excluded, creating a static model. The third alternative (denoted as Gated Recurrent Unit) is the dynamic computation network in Figure~\ref{fig:diagram} without system identification loss minimization. The fourth is the dynamic computation network without the data exchange. So, the fourth alternative trains the local agents independently of each other.
\begin{table*}[h]
\centering
\begin{sc}
\begin{tabular}{l|c|c|ccccccc}
\hline
\multirow{2}{*}{Methods}                              &\multirow{2}{*}{Metric}    &\multirow{2}{*}{AVG}      & \multicolumn{7}{c}{Locations}   \\
\cline{4-10}
                              &    &       & Albuquerque &Bakersfield &Boise &Casper &Denver &Las Vegas &Medford \\
\hline
\multirow{3}{*}{Logistic Regression} &\textbf{BCE}&3564.7 &2839.1   &2839.1      &2839.1 &5378.8&5378.7 &2839.1    &2839.1  \\
                                     & AUROC    & 50.0\%  &50.0\%   &50.0\%      &50.0\% &50.0\%&50.0\% &50.0\%    &50.0\%  \\
                                     & IOU      & 0.8 \%  &0.6\%    &0.9\%       &1.4\%  &0.5 \%&0.6\%  &0.4\%     &1.5\%   \\
\hline
\multirow{3}{*}{Generative Network}  &\textbf{BCE}& 33.7  &11.1     &35.8        &64.6  &6.7    &9.0    &3.2       &82.7    \\
                                     & AUROC    & 56.6\%  &74.9\%   &72.1\%      &54.2\%&50.0\% &50.7\% &49.9\%    &62.5\%  \\
                                     & IOU      & 0.9 \%  &1.8\%    &2.5\%       &1.7\% &0.0\%  &0.0\%  &0.0\%     &1.0\%   \\
\hline
\multirow{3}{*}{GRU}                 &\textbf{BCE}& 33.1  &16.9     &42.2        &64.7  &6.6    &8.9    &3.4       &89.0    \\
                                     & AUROC    & 54.3\%  &56.5\%   &61.8\%      &65.6\%&49.9\% &56.9\% &32.5\%    &56.9\%  \\
                                     & IOU      & 0.6 \%  &0.0\%    &4.4\%       &0.3\% &0.0\%  &0.0\%  &0.0\%     &0.7\%   \\
\hline
\underline{Proposed Method}          &\textbf{BCE}& 19.4  &9.3      &24.1        &28.6  &6.1    &7.3    &2.7       &57.5    \\
without                              & AUROC    & 69.6\%  &73.9\%   &76.9\%      &70.7\%&69.4\% &59.9\% &65.0\%    &71.2\%  \\
Sample Exchanges                     & IOU      & 1.9 \%  &1.7\%    &0.7\%       &1.3\% &4.8\%  &2.7\%  &0\%       &2.2\%   \\
\hline
\underline{Proposed Method}          &\textbf{BCE}& 14.3  &7.3      &17.5        &30.3  &3.8    &6.2    &1.7       &33.5    \\
using                                & AUROC    & 71.2\%  &71.5\%   &78.4\%      &73.4\%&70.6\% &59.3\% &77.3\%    &68.2\%  \\
\textbf{Sample Exchanges}            & IOU      & 1.2 \%  &1.3\%    &4.6\%       &1.5\% &0.0\%  &0.0\%  &0.0\%     &1.1\%   \\
\hline
\end{tabular}
\end{sc}
\caption{\small Performance of four weeks advance of fire grid map prediction. The evaluation of the models uses data from validation periods between 2012-05-02 and 2016-02-24. The train of the models uses the training time-series data span from  2000-11-01 to 2012-04-25.}
\label{tb:comparative_analysis}
\vskip -0.2in
\end{table*}

After 50 episodes of training, we evaluated the performance of the trained model. In each training episode, grid maps over the training period are sequentially fed into the model as time series. The evaluation result after the training, we evaluated their performance using the validation period as shown in Table~\ref{tb:comparative_analysis} where the performance metric values are time-averaged during the periods. From the top to the bottom row in Table~\ref{tb:comparative_analysis}, the model's complexity in terms of the number of components increases (so does the number of parameters of the model). In terms of the binary cross-entropy loss (BCE), the performance increases as the model complexity increase from the top to the bottom row.

The logistic regression in the first row of Table~\ref{tb:comparative_analysis} shows poor performance in terms of both BCE and Area Under ROC Curve (AUROC)
. AUROC 50\% of the logistic regression shows that the model does not have prediction capability. This poor performance was expected because the logistic regression uses a linear model that maps climate data at the pixel point in the grid map without considering the spatial correlation. The generative network in the second row of the table does consider the spatial correlation using convolution layers. As a result, the generative network improves BCE significantly compared to the logistic regression. The generative network is a static model similar to the previous uses of the static models in~\cite{jaafari2019hybrid, shang2020spatially, gholami2021there, coffield2019machine, mitsopoulos2017data}. So, the generative network does not consider temporal correlation.

In the third row of the table, we evaluate a model that can consider both spatial and temporal correlation. The model has the gated recurrent unit (GRU) in addition to the static components of the generative network. We denote the model as GRU as shown in Table~\ref{tb:comparative_analysis}. This GRU model is similar to the previous works~\cite{cheng2008integrated,huot2020deep,jin2020ufsp} that use a recurrent neural network such as GRU or \emph{Long Short Term Memory} (LSTM). In this model, the stream of the observations is dynamically processed by the GRU to predict the target fire grid map stream. The parameters of the model are trained by minimizing the prediction loss in~\eqref{eq:pred}. However, the GRU did not significantly improve the performance compared to the generative network in terms of BCE.

Now, we evaluate our proposed methods. The proposed method in the fourth row adds an objective, that is, system identification in addition to the prediction task. The observation predictor aims to predict the observations by minimizing the error between the predicted observations and the ground truth observations as embedded in the system identification loss function in~\eqref{eq:grad_sys_id}. In Figure~\ref{fig:one-step-prediction}, we can see that predicted observation grid maps correlate with the ground truth observations. This modification improved the performance in terms of BCE compared to the GRU. Finally, we evaluated the proposed method using the sample exchanges between local agents, governed by the policy trained by reinforcement learning. As shown in the last row of Table~\ref{tb:comparative_analysis}, the proposed method with \emph{sample exchanges} has improved BCE compared to the proposed method without sample exchanges.   

It is worth commenting on the performance in terms of the intersection of union (IOU) in Table~\ref{tb:comparative_analysis}. The IOU values of all the models in the table are very low and do not shows a meaningful trend in contrast to BCE or AUROC, which show improvements in the performances as the model complexities increase. Furthermore, as I commented earlier on Figure~\ref{fig:future-predictions}, the ground truth fire grid map doe not have clear segmentation as other benchmark data for segmentation tasks. Hence, IOU might not be a suitable performance metric for the fire grid map prediction.      

\section{Conclusion and Future Work}
In the past decade, wildfires have ravaged the western United States. Due to the scarcity of resources and difficulties in wildfire surveillance, first responders are often a step behind in fire response. In this research, we presented a novel, data-driven approach for extracting relevant fire determinants through readily available satellite images. We created an approach to wildfire prediction that explicitly takes state uncertainty into account. We also created a comprehensive model on wildfires that combines both historical fire data with relevant covarying data. Although there are some limitations to our current work, our forecasting model shows optimistic results and addresses the limitations in current wildfire prediction. Future works include more careful curation of the data that can be queried from the open-source remote sensing data repository by collaboration with domain experts in earth science. In addition, the data sharing between local agents need to be further studied since it is not clear how to leverage the data sharing for the fire prediction task toward optimality.

\bibliographystyle{IEEEtran}
\bibliography{mybib}


\section*{Appendix}

\begin{algorithm}[h]
   \caption{Two-Time-Scale Stochastic Optimization}
   \label{alg:multi-level-optimization}
\begin{algorithmic}
   \STATE {\bfseries Input:} Dynamic computational network in Figure~\ref{fig:diagram}
   \STATE {\bfseries \hspace{1cm}} with $\theta^\text{sys}_0$ of $\text{Enc}$, $\text{GRU}$, and $\text{Decoder}_0$.
   \STATE {\bfseries Input:} Fire map prediction network with $\theta^\text{pred}_0$ of $\text{Decoder}_0$.
   \STATE {\bfseries Input:} Data sharing reinforcement learning network 
   \STATE {\bfseries \hspace{1cm}} with $\theta^\text{critic}_0$ of $q(\cdot, \cdot)$ and $\theta^\text{actor}_0$ of $\mu(\cdot)$.
   \STATE {\bfseries Input:} Training time-series data.
   \STATE {\bfseries Input:} Replay buffers: $\mathcal{M}_\text{trajectory}$, $\mathcal{M}_\text{transition}$
   \STATE {\bfseries Output:} Fixed parameters: $\theta^\text{sys}_*$, $\theta^\text{pred}_*$, $\theta^\text{critic}_*$, and $\theta^\text{actor}_*$
   \REPEAT
   \FOR{$t=0$ {\bfseries to} $T_{\text{train}}$}
   \STATE For all local agents, update GRU state $\mathbf{h}_t$
   \STATE given $\mathbf{y}_t$ as in ~\eqref{eq:dynautoenc}
   \STATE $\quad \vect{h}_{t+1} \leftarrow \text{GRU}(\vect{h}_t, \text{Encoder}(\vect{x}_t))$
   \STATE $\quad \hat{\vect{x}}_{t} \leftarrow \text{Decoder}(\mathbf{h}_t)$.
   \STATE Add new data to the replay buffer
   \STATE $\quad \mathcal{M}_\text{trajectory} \leftarrow (\vect{x}_{t}, \vect{h}_{t})$.
   \STATE Sample a trajectory with length $T$ from $\mathcal{M}_\text{trajectory}$.
   \STATE Using the first time step of the sample trajectories from the agents, determine data sharing probability, i.e., $\vect{a}_n = \mu(\bar{\vect{h}}_n)$.
   \STATE Given $\vect{a}_n$, the sampled trajectories are assigned to each agent as illustrated in Figure~\ref{fig/network} and update parameters with the gradients in~\eqref{eq:grad_sys_id} and~\eqref{eq:pred}
   \STATE $\quad \theta^\text{sys}_{n+1}\leftarrow\theta^\text{sys}_{n}+\epsilon_n^\text{sys} S_n^\text{sys}(\mathcal{M}_\text{trajectory}, \theta^\text{actor}_{n})$
   \STATE $\quad \theta^\text{pred}_{n+1}\leftarrow\theta^\text{pred}_{n}+\epsilon_n^\text{pred} S_n^\text{pred}(\mathcal{M}_\text{trajectory}, \theta^\text{actor}_{n})$.
   \STATE Using the loss values obtained in calculating $S_n^\text{pred}(\mathcal{M}_\text{trajectory}, \theta^\text{actor}_{n})$ calculate rewards and add a transition data to the replay buffer
   \STATE $\quad \mathcal{M}_\text{transition} \leftarrow (\bar{\mathbf{h}}_{n}, \mathbf{a}_{n}, \bar{\mathbf{h}}_{n+1}, r_{n} )$.
   \STATE Update parameters with the gradients in~\eqref{eq:critic} and~\eqref{eq:grad_actor}
   \STATE $\quad \theta^\text{critic}_{n+1}=\theta^\text{critic}_{n}+\epsilon_n^\text{critic}
    S_n^\text{critic}(\mathcal{M}_\text{transition})$
   \STATE $\quad \theta^\text{actor}_{n+1} = \theta^\text{actor}_{n} +\epsilon_n^\text{actor} S_n^\text{actor}(\mathcal{M}_\text{transition})$
   \STATE Update the step sizes according to~\eqref{eq:step-size-rule}.
\ENDFOR
   \UNTIL{the performance meets the requirements.}
\STATE \textbf{Fix} parameters with the current ones.
\end{algorithmic}
\end{algorithm}

\subsection*{Some Details on Data Pre-Processing}
The binary target fire grid maps are generated using a threshold on the fire confidence value grid map from the FIRM dataset~\cite{davies2019nasa}. The range of the confidence is $[0,1]$ and the threshold that we chose to generate the binary target was set to $0.05$. The observation grid map as the input to the model has eleven channels, as shown in Figure~\ref{fig:one-step-prediction}. The pixel values of the eleven channels are normalized into the range of $[0,1]$. The grid maps covering the 200 km by 200 km square in Figure~\ref{fig/grid map} are resized into $64 \times 64$ grid maps. Time discretization of the daily time series data of FIRM and ERA5 Daily Aggregates~\cite{copernicus2017copernicus} uses the step size of a week by taking an average of climate data (ERA5) and the max over the week for FIRM.

\subsection*{Some Details on the Computational Neural Network}
The convolution layers in Figure~\ref{fig:diagram} that encodes the $64\times64$ grid map into a vector have four dimensional convolution layers and leaky rectifier leaky rectified linear units are placed between the convolution layers. Then, the vector from the convolution layers is fed to the gated recurrent unit (GRU) to update the state of the GRU. The state of the GRU is then feed to the deconvolution layers. The deconvolution layers in Figure~\ref{fig:diagram} implement the generator model in~\cite{radford2015unsupervised}. We use five set of deconvolution layers for the observation prediction and four future fire grid map predictions.

Training of the neural networks in the proposed framework is performed as multi-time scale stochastic optimization as in~\eqref{eq:step-size-rule}.  In~\cite{heusel2017gans}, it was shown that the \emph{ADAM}~\cite{kingma2014adam} step size rule can be set to implement the multi-time scale step size rule in~\eqref{eq:step-size-rule}. So, we used the \emph{ADAM} step size rule in~\cite{heusel2017gans}.

\end{document}